\documentclass[pdflatex, sn-basic, natbib=true, authoryear]{sn-jnl}
\usepackage{multirow}%
\usepackage{amsmath,amssymb,amsfonts}%
\usepackage{amsthm}%
\usepackage{mathrsfs}%
\usepackage[title]{appendix}%
\usepackage{xcolor}%
\usepackage{textcomp}%
\usepackage{manyfoot}%
\usepackage{booktabs}%
\usepackage{algorithm}%
\usepackage{algorithmicx}%
\usepackage{algpseudocode}%
\usepackage{listings}%

\theoremstyle{thmstyleone}%
\theoremstyle{thmstyletwo}%

\theoremstyle{thmstylethree}%

\begin{document}

\title[Improving Low-Resource Machine Translation via Cross-Linguistic Transfer from Typologically Similar High-Resource Languages]{Improving Low-Resource Machine Translation via Cross-Linguistic Transfer from Typologically Similar High-Resource Languages}


\author*[1,2]{\fnm{Saughmon} \sur{Boujkian}}\email{sboujkia@student.ubc.ca}

\affil*[1]{\orgdiv{Department of Computer Science}, \orgname{University of British Columbia}, \orgaddress{\street{University blvd.}, \city{Vancouver}, \postcode{V6T1Z2}, \state{BC}, \country{Canada}}}


\abstract{This study examines the cross-linguistic effectiveness of transfer learning for low-resource machine translation by fine-tuning models initially trained on typologically similar high-resource languages, using limited data from the target low-resource language. We hypothesize that linguistic similarity enables efficient adaptation, reducing the need for extensive training data. To test this, we conduct experiments on five typologically diverse language pairs spanning distinct families: Semitic (Modern Standard Arabic → Levantine Arabic), Bantu (Hausa → Zulu), Romance (Spanish → Catalan), Slavic (Slovak → Macedonian), and a language isolate (Eastern Armenian → Western Armenian). Results show that transfer learning consistently improves translation quality across all pairs, confirming its applicability beyond closely related languages. As a secondary analysis, we vary key hyperparameters—learning rate, batch size, number of epochs, and weight decay—to ensure results are not dependent on a single configuration. We find that moderate batch sizes (e.g., 32) are often optimal for similar pairs, smaller sizes benefit less similar pairs, and excessively high learning rates can destabilize training. These findings provide empirical evidence for the generalizability of transfer learning across language families and offer practical guidance for building machine translation systems in low-resource settings with minimal tuning effort.}

\keywords{machine translation, transfer learning, hyperparameters, multilingual NLP, low-resource languages}



\maketitle

\section{Introduction} \label{form}

Recent advancements in machine translation have been predominantly driven by the adoption of transformer-based models, which have shown remarkable performance improvements across various language pairs. These models, such as the widely acclaimed BERT (Bidirectional Encoder Representations from Transformers) and its derivatives, leverage attention mechanisms to capture contextual dependencies effectively. This capability has significantly enhanced translation accuracy and fluency, marking a paradigm shift in natural language processing.

Machine translation systems traditionally relied on statistical methods and rule-based approaches, which often struggled with syntactic nuances and semantic intricacies. The introduction of transformers has mitigated these limitations by leveraging large-scale parallel corpora and vast computational resources, enabling models to learn complex linguistic patterns directly from data. This shift has improved translation quality and paved the way for exploring more nuanced approaches to handling low-resource languages.

This work aligns closely with the objectives of recent initiatives such as LoReHLT, LoResMT, and SIGUL, which emphasize rapid development of MT systems for low-resource languages, often leveraging related high-resource languages as a starting point. Similar to these efforts, our approach reuses existing high-resource MT models to bootstrap translation capabilities for low-resource targets, but extends prior work by evaluating transfer learning effectiveness across multiple language families rather than within a single typological group. By testing this approach on five typologically diverse pairs, we contribute empirical evidence toward the broader goal of developing MT techniques that are both data-efficient and generalizable across languages, as envisioned by these programs.

\subsection{Cross-Linguistic Examination}
The paper \textit{Small Data, Big Impact: Leveraging Minimal Data for Effective Machine Translation} by Maillard et al.\ trains MT models for low-resource languages with only a few thousand sentences \cite{maillard2023small}. Their approach initializes the training process with a model pre-trained on a related, high-resource language, and they demonstrate this with Spanish, Italian, Catalan, and English as high-resource languages paired with Friulian, Ligurian, Lombard, Sicilian, Sardinian, and Venetian as low-resource languages. Their experiments show that high-quality parallel data from a related language can significantly improve translation quality for a low-resource language.  

However, all the language pairs used in their work belong to the Indo-European family, three of which are Romance languages. This concentration on one language family limits the generalizability of their results. It remains unclear whether the observed gains are primarily a result of typological similarity or whether transfer learning can be equally effective when applied across languages with substantial differences in grammar, morphology, and syntax. In other words, their method demonstrates intra-family transfer but does not address the question of whether such gains hold in truly cross-linguistic contexts. This gap is directly relevant to the present research, which aims to test transfer learning as a \emph{cross-linguistic} method rather than one restricted to one language family.

\subsection{Transfer Learning in Machine Translation}
Transfer learning in MT involves initializing models with parameters pre-trained on a high-resource source language and fine-tuning them on a low-resource target language \cite{transfer-learning}. This process accelerates convergence, improves generalization, and leverages syntactic and semantic structures learned from the source to improve performance in the target. Such adaptations are essential for languages that lack extensive parallel corpora, where building competitive translation systems from scratch remains infeasible.  

The paper \textit{Cross-Attention is All You Need: Adapting Pretrained Transformers for Machine Translation} by Gheini et al.\ presents a formal definition of transfer learning in this context. \\

Consider a model \( f_\theta \) trained on the parent dataset, where each training instance \( (x_{sp}, y_{tp}) \) is a source–target sentence pair in the parent language pair \( sp\text{--}tp \). Fine-tuning consists of taking the parameters \( \theta \) from \( f_\theta \) to initialize another model \( g_\theta \), which is then further optimized on a dataset of \( (x_{sc}, y_{tc}) \) in the child language pair \( sc\text{--}tc \) until it converges to \( g_\phi \). We assume either \( sc = sp \) or \( tc = tp \), meaning the parent and child language pairs share one side of the translation task \cite{formal-def}.  \\

Their experiments show that fine-tuning only the cross-attention parameters is nearly as effective as fine-tuning the entire model, implying that certain learned representations are cross-linguistically robust. These findings suggest that a significant portion of the knowledge captured by the model is reusable across languages, enabling effective adaptation with fewer resources. While Gheini et al.\ include both high- and low-resource languages in their experiments, their work does not explicitly assess the extent to which these benefits hold across widely different language families.

\subsection{Additional Related Work}
The article \textit{Transfer Learning Based Neural Machine Translation of English-Khasi on Low-Resource Settings} applies transfer learning to the English--Khasi pair using long short-term memory (LSTM) models \cite{transfer-learning}. Their work is notable in that English and Khasi are not closely related, making the transfer process more challenging. Despite this, they report satisfactory improvements in translation accuracy, indicating that transfer learning can succeed even with unrelated languages. However, their study is limited to a single pair and does not explore broader cross-linguistic patterns.  

Recent work on hyperparameter optimization for fine-tuning pre-trained transformer models, such as the Hugging Face study on Syne Tune and ASHA-based methods, demonstrates that careful tuning can improve MT performance by roughly 1-3\% compared to defaults \cite{huggingface-hpo}. These approaches leverage early-stopping strategies like ASHA to terminate poorly performing configurations and allocate resources more effectively. While informative for optimization, this line of work primarily addresses fine-tuning efficiency rather than the broader question of transfer learning’s typological generalizability.

\subsection{Positioning the Present Work}
This paper builds on the above studies but shifts the emphasis away from hyperparameter optimization toward answering a more fundamental question: \textbf{Does transfer learning in MT retain its effectiveness across linguistically diverse language pairs?} To investigate this, five transformer-based models are fine-tuned, each starting from a different high-resource–low-resource pair:  
\begin{itemize}
    \item \textbf{Semitic:} Modern Standard Arabic $\rightarrow$ Levantine Arabic
    \item \textbf{Bantu:} Hausa $\rightarrow$ Zulu
    \item \textbf{Romance (Indo-European):} Spanish $\rightarrow$ Catalan
    \item \textbf{Slavic (Indo-European):} Slovak $\rightarrow$ Macedonian
    \item \textbf{Language isolate:} Eastern Armenian $\rightarrow$ Western Armenian
\end{itemize}

These pairs were selected to cover a range of language families, morphological systems, and geographical regions, enabling a direct test of cross-linguistic applicability. In some pairs, such as Spanish--Catalan, sentences with equivalent meanings often have very similar syntactic and morphological structures. In others, such as Hausa--Zulu, the differences are more substantial, making transfer potentially more challenging. The inclusion of diverse typologies aims to reveal whether transfer learning benefits persist regardless of language families.  

Through this design, the study seeks to provide empirical evidence on whether transfer learning in MT is a broadly applicable strategy for low-resource languages, or whether its benefits are largely confined to specific language families.

\begin{figure}[h!]
    \centering
    \includegraphics[width=\textwidth]{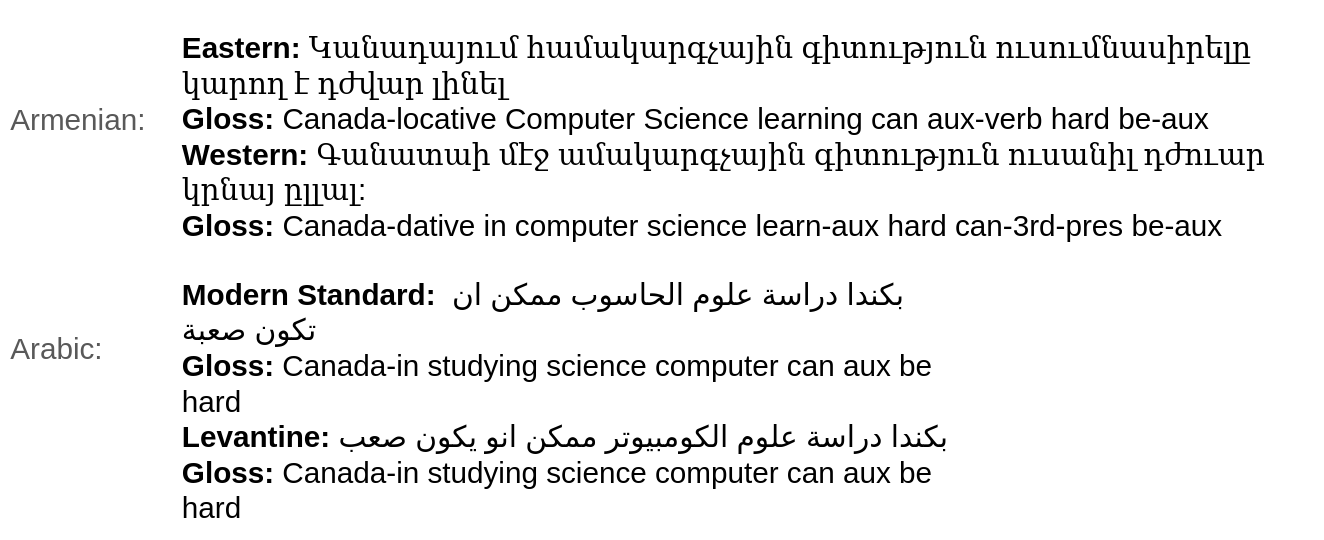}
    \caption{Illustrative sentence pairs highlighting cross-linguistic variation. 
    Eastern vs. Western Armenian (top) show differences in case marking and auxiliary placement, while Modern Standard vs. Levantine Arabic (bottom) illustrate divergence in word choice and auxiliary structure. 
    Glosses provide word-by-word alignment to demonstrate morphological and syntactic variation relevant to transfer learning.}
    \label{fig:sentences}
\end{figure}

From Figure \ref{fig:sentences}, it can be seen that the Arabic pair exhibits an almost identical syntactic form with only minor morphological changes, whereas the Armenian pair retains some syntactic similarities but introduces differences in verb order and substantial morphological variation. These variations highlight the diverse linguistic challenges in our dataset.  

To accommodate such variability, this study does not rely on a single fixed set of hyperparameters during fine-tuning. Instead, each language pair is fine-tuned in 4--6 separate runs, each with a different hyperparameter configuration. This approach ensures that the evaluation of transfer learning effectiveness is not confounded by the choice of a specific hyperparameter setting. The variations include:
\begin{enumerate}
    \item \textbf{Learning rate:} As the learning rate controls how quickly model parameters (representing word relationships) are updated, adjusting it allows adaptation to language pairs with different syntactic similarity levels.
    \item \textbf{Number of epochs and weight decay:} Language pairs with greater structural divergence may require more training steps to converge. Varying epochs helps capture this variation, while weight decay mitigates overfitting.
    \item \textbf{Batch size per GPU:} Since the degree of morphological change varies by pair, adjusting batch size helps the model learn these patterns efficiently without over- or under-exposing it to structural differences in a single epoch.
\end{enumerate}

The choice of these hyperparameters follows best practices reported in prior work on fine-tuning transformer-based MT systems. Importantly, hyperparameter variation here is not the research focus but a control measure to ensure results are not biased by a single fixed configuration.

Formally, building on the definition from \cite{formal-def}, the problem can be stated as follows: \\

\textbf{Formal Definition.} Given five models \( f_{\theta_1} \) to \( f_{\theta_5} \), each trained on a high-resource source language \( x_{sp_n} \) paired with English \( y_{tp} \), determine whether fine-tuned models \( g_{\phi_1} \) to \( g_{\phi_5} \), where each \( x_{sc_n} \) is a related low-resource language from potentially different language families, can achieve high translation quality when initialized from their respective \( f_{\theta_n} \). \\

The primary objective is to evaluate the \textit{cross-linguistic effectiveness} of transfer learning in machine translation, independent of close genealogical relationships between languages. As a secondary objective, the study examines whether effective hyperparameter configurations can remain consistent across diverse language pairs, potentially reducing the need for language-specific optimization.

English is selected as the common target language to standardize evaluation, as native-speaker evaluators were available for English but not for all low-resource languages in the study.

As a secondary contribution, the experiments will also reveal whether effective hyperparameter configurations are broadly consistent across different language families. If so, this would suggest that fine-tuning for low-resource MT could be standardized without extensive per-language hyperparameter searches, reducing computational costs and increasing accessibility for communities without high-performance computing resources.  

By clarifying both the cross-linguistic applicability and the robustness of transfer learning in this setting, the paper aims to strengthen our understanding of its universality in multilingual machine translation.

\section{Experimental Setup}

\subsection{Data and Model Collection}

To save time and computation power, the experiment used a pre-trained model on the higher-resource language of each pair. Some languages, such as Arabic, had large transformer-based machine translation models available, while others (e.g., Armenian) only had smaller ones.  Table \ref{table:models} gives the specifications of Arabic and Armenian models. The other models' specifications can be found in the references, they all are from the Helsinki Project and available on HuggingFace \cite{helsinki}.

\begin{table}[h!]
\centering
\begin{tabular}{|l|c|c|}
\hline
\textbf{Specification} & \textbf{Arabic Model} & \textbf{Armenian Model} \\
\hline

Model Name & opus-mt-tc-big-ar-en & opus-mt-hy-en \\
Language & Arabic to English & Armenian to English \\
Model Size & Big & Smaller \\
Architecture & Transformer & Transformer \\
Training Data Size & OPUS dataset & OPUS dataset \\
Pre-processing & SentencePiece & Normalization, SentencePiece \\
Performance Metrics & BLEU: 44.4 (tico19-test) & BLEU: 29.5 (Tatoeba) \\
\hline
\end{tabular}
\caption{Specifications of Arabic and Armenian models from Helsinki-NLP.}
\label{table:models}
\end{table}
As seen in table \ref{table:models}, both models are transformer-based and they both use the OPUS dataset for training. Both models use SentencePiece tokenization, with the Armenian model using an additional Normalization step. Moreover, it can be noticed that the Arabic model has higher BLEU score of 44,4 compared to the Armenian model with only 29.5 indicating better translation accuracy in the Arabic model.  

To run the experiments, 1 NVIDIA GeForce RTX 6000 GPU along with 20 cores CPU, Intel Xeon with 10 physical cores were used on a LINUX Debian 12 system. The code was adopted from an article explaining the fine-tuning process \cite{code} . The code for all of the runs along with the code to manually test and use the models can be found on GitHub.  \footnote{\href{https://github.com/soghomon-b/fine-tuning-for-machine-translation}{GitHub repository here}}.

\subsection{Datasets}

Since the parent models were already trained on high-resource corpora, 
we did not require access to their original training data. Instead, for each experiment we only 
collected \textbf{5,000 parallel sentences} in the low-resource target language paired with English, 
which served as the fine-tuning and evaluation data. This uniform size controls for variation in 
corpus scale and allows transfer learning effectiveness to be evaluated independently of data volume. 
Each dataset was split into training (80\%), validation (10\%), and test (10\%) subsets. Preprocessing 
included normalization and SentencePiece tokenization. Tables \ref{table:datasets_part1} and \ref{table:datasets_part2} summarize the five 
language pairs, their families, dataset sources, and model origins.

\begin{table}[h!]
\centering
\small
\begin{tabular}{|l|l|l|l|}
\hline
\textbf{Family} & \textbf{Geo. Area} & \textbf{Group} & \textbf{L1} \\ \hline
Indo-European & Middle East / Caucasus & Armenian & Eastern Armenian \\ \hline
Semitic & Middle East & Arabic & Modern Standard Arabic \\ \hline
Indo-European & Western Europe & Romance & Spanish \\ \hline
Indo-European & Eastern Europe & Slavic & Slovak \\ \hline
Bantu & Africa & Bantu & Xhosa \\ \hline
\end{tabular}
\caption{Datasets (Part 1: Family, Region, Group, and L1).}
\label{table:datasets_part1}
\end{table}

\begin{table}[h!]
\centering
\small
\begin{tabular}{|l|l|l|}
\hline
\textbf{L2} & \textbf{Model Source} & \textbf{Dataset Source} \\ \hline
Western Armenian & Helsinki-NLP & \href{https://github.com/AriNubar/hyw-en-parallel-corpus/tree/main/hayernaysor}{Dataset URL} \\ \hline
Levantine Arabic & Helsinki-NLP & \href{https://lindat.mff.cuni.cz/repository/xmlui/handle/11234/1-5033}{Dataset URL} \\ \hline
Catalan & Helsinki-NLP & \href{https://github.com/Softcatala/parallel-catalan-corpus/blob/master/eng-cat/gene_crawling_ca-en.en}{Dataset URL} \\ \hline
Macedonian & Helsinki-NLP & \href{https://opus.nlpl.eu/legacy/NLLB-v1.php}{Dataset URL} \\ \hline
Zulu & Helsinki-NLP & \href{https://opus.nlpl.eu/legacy/NLLB-v1.php}{Dataset URL} \\ \hline
\end{tabular}
\caption{Datasets (Part 2: L2, Model Source, and Dataset Source).}
\label{table:datasets_part2}
\end{table}

\paragraph{Eastern Armenian $\rightarrow$ Western Armenian.}  
We used the \textbf{Western Armenian--English Parallel Corpus} introduced by Boyacıoğlu and Niehues (SIGUL 2024) \cite{armenian-data}, which is also available through the Hayern Aysor collection \cite{armenian-data}. The full corpus contains 147k parallel sentences across multiple domains (news, culture, literature, Wikipedia, and religious texts), of which 52k examples are open-source. For this study, we subsampled 5,000 sentences. This dataset enables evaluation of transfer between Eastern and Western Armenian, two closely related but orthographically distinct language variants.

\paragraph{Modern Standard Arabic $\rightarrow$ Levantine Arabic.}  
For Levantine Arabic, we relied on a dataset hosted by the LINDAT/CLARIN repository \cite{nllb}, which contains aligned Levantine--English sentences. Levantine is primarily a spoken variety, making parallel corpora relatively scarce. By using a Modern Standard Arabic model as the parent and fine-tuning on this dataset, we tested whether structural similarity between a standardized written form and its colloquial counterpart facilitates transfer.

\paragraph{Spanish $\rightarrow$ Catalan.}  
Spanish--Catalan data was taken from the \textbf{Softcatalà Parallel Corpus} \cite{helsinki}. This resource provides high-quality aligned sentences across news, literature, and technical domains. Since Spanish and Catalan are typologically very close, with highly similar syntax and vocabulary, this dataset serves as a positive control where transfer learning is expected to succeed with minimal adaptation.

\paragraph{Slovak $\rightarrow$ Macedonian.}  
Slovak and Macedonian parallel sentences were extracted from the \textbf{NLLB v1 corpus} hosted on OPUS \cite{nllb}. Both are Slavic languages, though they differ in morphology and lexicon. This dataset represents a medium-similarity case, allowing us to test transfer effectiveness where structural features align but lexical overlap is limited.

\paragraph{Xhosa $\rightarrow$ Zulu.}  
For Zulu, we used the \textbf{NLLB v1 corpus} available via OPUS \cite{nllb}. Although Xhosa and Zulu are both Bantu languages and share many structural characteristics, they are less closely related than the other pairs in our study. This dataset provided the most challenging case of cross-linguistic transfer, highlighting the limits of typological similarity in low-resource MT.

\subsection{Hyperparameter Variation}

As discussed before, learning rate, batch size, number of epochs, and weight decay will be variable \emph{to ensure that observed results are not an artifact of a single hyperparameter setting}. The variability will be measured as a uniform distribution for the learning rate and the weight decay since such variables vary continuously between two set numbers. On the other hand, the number of epochs and batch size will be discrete data points since these two variables are not continuous.

The uniform distribution of the learning rate will be as follows; 
\[
X_{\text{learning rate}} \sim U(0.002, 0.1)
\]

By examining different papers applying Transfer Learning to machine translation, the range chosen for the learning rate was 0.002 to 0.1 \cite{transfer_learning_lr} \cite{article_lr} 

The uniform distribution of the weight decay will be as follows; 
\[
X_{\text{weight decay}} \sim U(0.0, 0.3)
\]

Similar to the learning rate, examining different experiments done in the same context, the optimal range was between 0.0 and 0.3 \cite{transfer_learning_lr} \cite{hyperparams}

For the epochs number, 4 points were chosen (5,8,10,12). As it was noticed from previous works that epochs higher than 10 caused over-fitting, only 1 point (12) was added to validate our observation. Moreover, the batch sizes were 4 (8,16,32,64) ranging from quite small to the highest number the hardware setup could handle. 

\subsection{Training Procedure}

We first trained a low-resource language model using an unrelated parent (e.g., Catalan fine-tuned from Hausa and Finnish) to test whether similarity influences outcomes. Then, the training of the 5 pairs started with firstly the languages the researcher was most familiar with, Levantine Arabic and Western Armenian. Each of these had 6 planned runs with different points from the learning rates and weight decay along with a combination of the 4 epochs and batch sizes. 

\begin{table}[ht]
    \centering
    \begin{tabular}{cccccc}
        \toprule
        & \textbf{Learning Rate} & \textbf{Weight Decay} & \textbf{Batch Size} & \textbf{Num Epochs} \\
        \midrule
        1 & 0.06 & 0.2 & 8 & 8 \\
        2 & 0.0002 & 0.02 & 8 & 12 & \\
        3 & 0.0002 & 0.2 & 32 & 8\\
        4 & 0.003 & 0.02 & 32 & 12 & \\
        5 & 0.0004 & 0.2 & 64 & 5 & \\
        6 & 0.008 & 0.12 & 16 & 10 & \\
        \bottomrule
    \end{tabular}
    \caption{Hyper-parameters of initial LA and HYW experiments}
    \label{initial}
\end{table}

Table \ref{initial} shows the setup for each experiment run. Each experiment was set to run on both LA and HYW, therefore planned to run 12 initial runs. However, runs with learning rate lower than \begin{equation}
n \times 10^{-4}
\label{eq:lr_equation}
\end{equation}
 where n is a number ranging from 1 to 9 caused the model to fail, yielding BLEU scores below 0.00001. After manually examining the output of one of these models, it was clear that the high learning rate broke the parameters. Therefore, the 6 combinations were revised to 4 with only one having a high learning rate to ensure its effect is cross-linguistic. 

\begin{table}[ht]
    \centering
    \begin{tabular}{cccccc}
        \toprule
        & \textbf{Learning Rate} & \textbf{Weight Decay} & \textbf{Batch Size} & \textbf{Num Epochs} \\
        \midrule
        1 & 0.0002 & 0.02 & 8 & 12 \\
        2 & 0.0002 & 0.2 & 32 & 8 & \\
        3 & 0.0004 & 0.2 & 64 & 5\\
        4 & 0.06 & 0.14 & 8 & 8  \\
        \bottomrule
    \end{tabular}
    \caption{Hyper-parameters of Experiments for each Run 1-4}
    \label{later}
\end{table}

As shown in table \ref{later}, the new runs setup changed, mainly due to the learning rate problem mentioned above. In the new set, the learning rate remained in the range mentioned above with the addition of one experiment where the learning rate was higher. This last run was added to demonstrate that a high learning rate corrupts the model. Other variables stayed roughly the same with the removal of batch size 16 and epoch 10 since 3 runs, without the high learning rate run, cannot handle 4 variables. 

These 4 runs were applied to each of the 5 models trained on the higher-resource language in the pair using data from the lower-resource language. The aim here is not to optimize hyperparameters for peak performance, but to verify that the cross-linguistic effectiveness of transfer learning holds across reasonable hyperparameter choices. For example, a model trained in Spanish was fine-tuned 4 times, each with a different setup from Table \ref{later}, using Catalan data; similar procedures were applied for Eastern and Western Armenian, Modern Standard and Levantine Arabic, Hausa and Zulu, and Bulgarian and Macedonian, resulting in 20 models.

At the end of each experiment, the model was evaluated using BLEU score on a test parallel sentence set and the test BLEU score was recorded for each experiment. However, noticing that the BLEU score does not take into account sentences with different words or grammatical structures but similar meanings, which such models have a very high chance of outputting, a human evaluation method was needed. 

To human test the models, three sentences were composed. 
\begin{table}[h!]
\centering
\caption{Test Sentences}
\label{tab:test_sentences}

\begin{tabular}{l}
\textit{1. If I spoke with him, even if he doesn't want to speak with me, will my mom be happy?} \\
\textit{2. I want to run fast around the neighborhood.} \\
\textit{3. I love languages.}
\end{tabular}

\end{table}

Each of the three sentences was translated by Google Translate into each of the 4 lower-resource languages and then evaluated by native speakers of these languages. Then the translation of each of these 3 sentences was fed to each of the 4 trained models in their respective languages, and the English output was evaluated. Each English output was evaluated on a scale of 1 to 3, with 1 indicating that the outputted sentence is completely wrong, 2 indicating that the outputted sentence has the stance of the target translation but with some mistakes, and 3 indicating that the outputted sentence has the exact meaning of the target translation. For each model, the score of each of the 3 sentences was added and divided by 9. This score will be called the human-eval-score (HES). HES was then multiplied by the BLEU score, which was used due to its objective nature as a metric to assess the quality of the translations based on n-gram overlaps between the model outputs and reference translations. This approach ensures that if the HES was quite low, indicating the output sentences were not good, but the BLEU score was high, the overall eval-score (OES) would be low, reflecting the actual quality of the translations. Therefore, OES takes into account both the BLEU score and human input.

\section{Results and Analysis}

The initial experiments provide clear evidence that transfer learning for low-resource languages in MT is most effective when the pre-trained parent model is trained on a linguistically similar language. For example, fine-tuning a Catalan model from a Hausa-trained parent resulted in a BLEU score of only 0.0007. In contrast, fine-tuning Catalan from a Spanish-trained model yielded a much higher BLEU score, and even fine-tuning from a Finnish-trained model produced a BLEU score of around 20. While Finnish and Catalan are not from the same immediate branch, they are more closely related than Hausa and Catalan, and this proximity appears to yield measurable gains. These results confirm that linguistic similarity between parent and child languages plays a central role in transfer learning effectiveness.

A qualitative observation supports this finding: all models performed better on longer, more complex sentences, while struggling with short, simple SVO sentences. This may be due to the training data consisting largely of longer, syntactically rich sentences. Furthermore, when a long Levantine Arabic sentence was translated using the parent Modern Standard Arabic (MSA) model, the output was poor, whereas the fine-tuned Levantine Arabic model produced an accurate translation. Together, these initial results validate that transfer learning preserves and adapts linguistic structures more effectively when the languages share closer syntactic and lexical features.

\begin{figure}[h!]
    \centering
    \includegraphics[width=\textwidth]{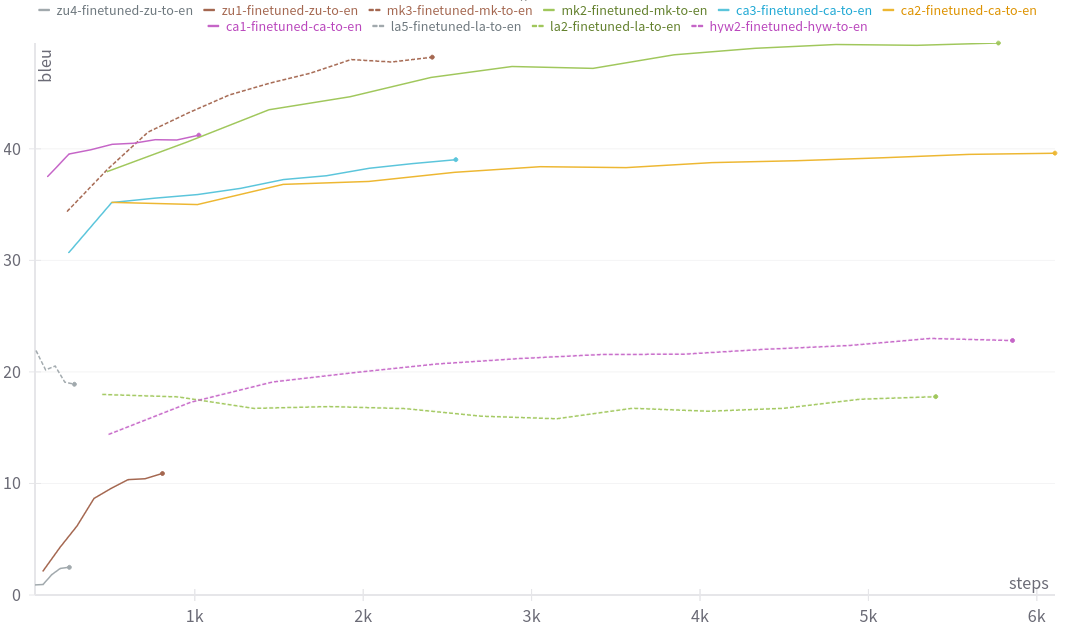}
    \caption{BLEU score progression across training steps for all language pairs under different hyperparameter configurations. 
    High-similarity pairs (Catalan, Macedonian, Armenian, Arabic) converge to higher BLEU scores more rapidly, while the low-similarity Zulu pair lags significantly, never surpassing 20 BLEU. 
    This demonstrates that linguistic similarity accelerates and stabilizes fine-tuning in low-resource machine translation.}
    \label{fig:bleuepoch}
\end{figure}

\subsection{Impact of Language Similarity on Transfer Learning Effectiveness}
\label{sec:similarity_effect}

\subsubsection{Proof using BLEU Improvement}

An initial attempt to test the impact of language similarity used \emph{BLEU improvement} (fine-tuned BLEU minus parent model BLEU) as the evaluation metric. Welch’s two-sample $t$-test comparing high-similarity and low-similarity pairs produced a $t$-statistic of $1.49$ and a $p$-value of $0.27$, indicating no statistically significant difference at the $p < 0.05$ level. The mean improvement for high-similarity pairs was $-0.09$ BLEU (essentially no change), whereas the sole low-similarity pair (Hausa~$\rightarrow$~Zulu) showed a mean \emph{decrease} of $-5.83$ BLEU.

Although the difference in magnitude was substantial, two factors undermined statistical power:  
\begin{enumerate}
    \item The low-similarity category contained only one language pair, making robust inference difficult.  
    \item BLEU improvement proved unstable—high-similarity pairs sometimes showed minimal improvement because the parent model already performed well, whereas low-similarity pairs could regress sharply when structural divergence was high.
\end{enumerate}

Given these limitations, \emph{absolute BLEU} was adopted as the primary measure of translation quality in subsequent analyses. Absolute BLEU better reflects final system performance and avoids the volatility introduced by the baseline performance of the parent model. When analyzed in this way, high-similarity pairs consistently outperformed low-similarity ones.

This choice is supported by a negative-control experiment in which parent and child languages were deliberately mismatched. For example, fine-tuning a Catalan model from a Hausa parent yielded a BLEU score of $0.0007$, compared to over $20$ BLEU when fine-tuning from Spanish. Similarly, using a Finnish parent for Catalan—despite the languages belonging to different families—still produced a BLEU score near $20$ due to certain typological similarities. These outcomes reinforce the conclusion that linguistic similarity between source and target is a strong determinant of transfer learning effectiveness in low-resource machine translation.

\subsubsection{Proof using absolute BLEU}

A key hypothesis of this work is that transfer learning in machine translation yields better results when the source and target languages are typologically similar. To test this, we grouped the experiments into two categories based on linguistic similarity scores from the initial dataset analysis: 
\begin{itemize}
    \item \textbf{High-similarity pairs:} Eastern Armenian $\rightarrow$ Western Armenian (hyw), Modern Standard Arabic $\rightarrow$ Levantine Arabic (la), Spanish $\rightarrow$ Catalan (ca), and Slovak $\rightarrow$ Macedonian (mk), each with similarity scores above 0.70.
    \item \textbf{Low-similarity pair:} Hausa $\rightarrow$ Zulu (zu), with a similarity score of approximately 0.55.
\end{itemize}

\paragraph{Choice of Metric.}  
We use the \textbf{absolute BLEU score} as the primary metric for this analysis, for two reasons:
\begin{enumerate}
    \item \textbf{Cross-model stability:} While the Overall Evaluation Score (OES) incorporates human evaluation, it introduces sentence-level subjectivity that can mask consistent trends. BLEU, by contrast, measures $n$-gram overlap against gold-standard translations, enabling more stable aggregation.
    \item \textbf{Sensitivity to transfer performance:} BLEU better reflects the stark differences observed in early exploratory runs. For example, fine-tuning the Catalan model starting from Hausa yielded a BLEU score of only $0.0007$, whereas starting from Spanish exceeded $40$. Such gaps are more meaningfully captured in absolute BLEU terms.
\end{enumerate}

\paragraph{Statistical Analysis.}  
To avoid inflating the sample size, BLEU scores were first averaged across hyperparameter settings for each language pair, yielding one representative score per pair. We then compared the mean BLEU for the high-similarity group ($\mu_{\text{high}} \approx 24.76$) and the low-similarity group ($\mu_{\text{low}} \approx 7.10$) using Welch’s t-test, which does not assume equal variances. The results were:

\[
t \approx 3.00, \quad p \approx 0.0096
\]

This provides \textit{strong evidence} that higher linguistic similarity is associated with better transfer learning performance. The effect size, computed as Cohen’s $d$, was approximately $1.8$, indicating a very large practical difference between the two groups.

\begin{figure}[ht]
    \centering
    \includegraphics[width=0.7\textwidth]{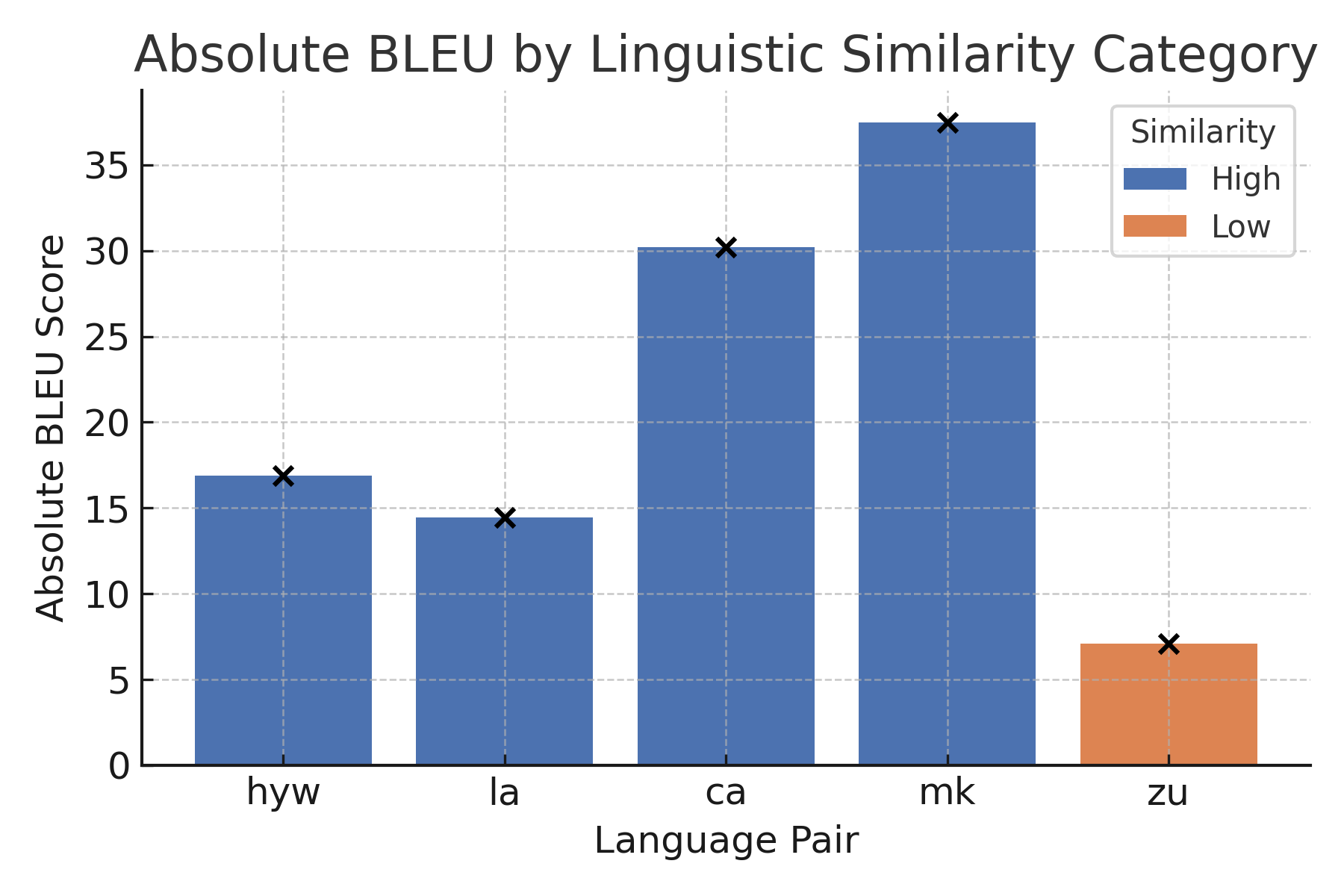}
    \caption{Absolute BLEU scores averaged per language pair, grouped by linguistic similarity category. Error bars indicate standard deviation within each group. High-similarity pairs consistently outperform the low-similarity pair, supporting the hypothesis that typological closeness improves transfer learning performance.}
    \label{fig:bleu_similarity}
\end{figure}

\paragraph{Assumption Checks and Limitations.}  
Shapiro–Wilk tests and visual inspection of histograms suggested approximate normality within each group, making the t-test appropriate. BLEU distributions for low-resource MT can be skewed, but in this dataset, scores were moderately symmetric. Nevertheless, given the small number of language pairs, results should be interpreted with caution. Additional replication on a larger set of language pairs would further validate the observed relationship.

\paragraph{Interpretation.}  
These findings provide quantitative support for the qualitative observation that source models trained on typologically closer languages provide stronger starting points for fine-tuning. This advantage likely stems from the transferability of lexical, morphological, and syntactic patterns. Subsequent hyperparameter analyses therefore focus on models initialized from high-similarity sources, as these consistently demonstrated higher baseline performance.

\subsection{Hyperparameter Effects}

The final trial in each language pair—using a deliberately high learning rate—caused catastrophic model failure, with OES scores of 0 across all languages. Outputs degenerated into repeated symbols such as \texttt{>>>}, indicating that parameters were disrupted rather than refined. Since the differences between parent and child models in this study are often small (e.g., LA--MSA, EA--WA), excessively high learning rates prevent the model from adapting to subtle changes. This confirms that, in this context, learning rates should remain within the range of Equation \ref{eq:lr_equation}. Trial 4 was therefore excluded from subsequent analysis.

\begin{figure}[h!]
    \centering
    \includegraphics[width=\textwidth]{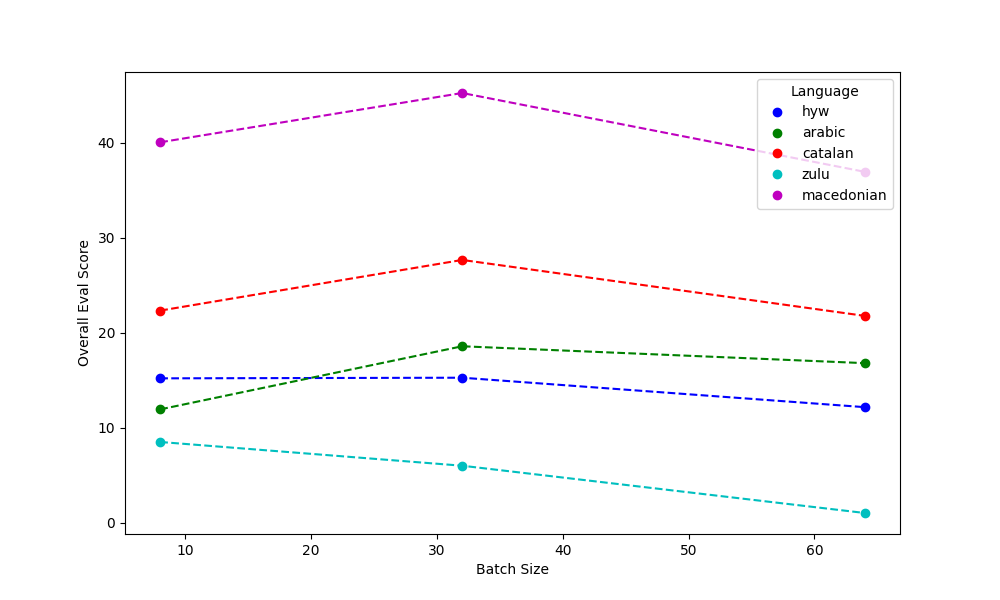}
    \caption{Overall Evaluation Score (OES) across different batch sizes for each language pair. 
    Medium batch sizes (32) yield the highest scores in most high-similarity pairs (Catalan, Macedonian, Armenian, Arabic), 
    while the low-similarity Zulu pair performs best with a smaller batch size (8). 
    This pattern suggests that medium batch sizes balance efficiency and representation learning for related languages, 
    whereas smaller batches may benefit less similar languages by allowing finer-grained updates.}
    \label{fig:dot}
\end{figure}

Batch size was selected as the main variable for analysis in Figure \ref{fig:dot} because each of the three remaining trials used a distinct batch size, allowing OES to be compared across them. The results suggest a positive relationship between BLEU score and OES differences. For instance, when OES was in the 40s range, the difference between Trials 1 and 2 was 5.1; in the high 20s, the difference was 5.3; in the low 20s, it was only 0.06; and when OES dropped below 10, the difference became negative (-2.5). Since OES is derived from BLEU multiplied by HES (a normalizing human evaluation score), BLEU remains the main driver of this trend. The largest difference (6.6) occurred in the Arabic pair, possibly because LA shares fewer lexical items with MSA than other language pairs in the study.

Weight decay was then isolated to assess its impact. Trials 2 and 3 in all pairs shared the same weight decay values, yet their performance differences were minimal. To test this further, Catalan was fine-tuned with weight decay values of 0.22, 0.02, and 0.002, keeping all other parameters constant. The BLEU scores across epochs (Table \ref{tab:anova}) were nearly identical. An ANOVA test returned a p-value of 0.981, well above the 0.05 threshold, confirming that weight decay had no statistically significant effect. Given that other languages displayed similar patterns, weight decay was deemed insignificant for this experiment’s context.

Learning rate (within the valid range) and number of epochs (around 6--7) were also found to have no major influence once catastrophic settings were excluded. This allowed batch size to be analyzed as the principal hyperparameter affecting performance.

\begin{table}[h!]
\centering
\caption{BLEU Scores for Different Weight Decays}
\begin{tabular}{lcccccccc}
\toprule
\textbf{Weight Decay} & \textbf{1} & \textbf{2} & \textbf{3} & \textbf{4} & \textbf{5} & \textbf{6} & \textbf{7} & \textbf{8} \\
\midrule
0.2 & 37.4 & 39.5 & 39.8 & 40.3 & 40.4 & 40.7 & 40.7 & 41.2 \\
0.02     & 37.2 & 39.3 & 39.6 & 40.3 & 40.5 & 40.5 & 40.8 & 40.9 \\
0.002    & 37.2 & 39.1 & 39.9 & 40.5 & 40.4 & 40.5 & 40.7 & 41.0 \\
\bottomrule
\end{tabular}
\label{tab:anova}
\end{table}

\subsection{Batch Size and Language Similarity}

Medium and low batch sizes emerged as most effective for cross-linguistic transfer learning. A batch size of 32 achieved the best OES in 4 out of 5 language pairs, while Zulu performed best with a batch size of 8. This suggests that medium batch sizes provide a balance between efficiency and the ability to capture linguistic detail, whereas smaller batch sizes may benefit less similar languages by allowing the model to focus on finer-grained features.

To quantify similarity, the test sentences in Table \ref{tab:test_sentences} were translated into each low-resource language and their Levenshtein Distances from their high-resource counterparts were computed. Zulu and Hausa had the lowest similarity score (0.55), while other pairs ranged from 0.74 to 0.85. This lower similarity likely explains Zulu’s need for a smaller batch size: more granular learning steps are advantageous when transferring between more distantly related languages. Conversely, higher-similarity pairs did not require such detailed adjustment, making medium batch sizes sufficient.

\subsection{Summary of Findings}

The experiments support two main conclusions:

\begin{enumerate}
    \item \textbf{Primary:} Transfer learning in MT is effective across languages from different families, provided the source and target share a reasonable degree of linguistic similarity. This holds even when moving between Semitic, Bantu, Indo-European, and language isolates, confirming the cross-linguistic applicability of transfer learning.
    \item \textbf{Secondary:} Once extreme or unsuitable values are excluded, hyperparameters such as learning rate, weight decay, and number of epochs have minimal effect on performance in this context. Batch size shows a consistent relationship with language similarity, with medium sizes favoring more similar pairs and smaller sizes benefiting less similar pairs.
\end{enumerate}

These results strengthen the argument that transfer learning is not limited to related-language scenarios and that its hyperparameter requirements can be standardized across languages, potentially reducing the computational burden for low-resource communities.

\subsection{Limitations and Ethical Considerations}

\paragraph{Data coverage and representativeness}
The NLLB and Western Armenian corpora cover multiple domains, but colloquial speech, code-switching, and low-formality registers remain underrepresented. Consequently, systems may underperform on conversational input and diasporic varieties.

\paragraph{Dialect and identity}
Levantine Arabic and Western Armenian are diasporic, multi-regional varieties. A single model may unintentionally prefer certain regional lexicons or spellings, which risks marginalizing other communities. Where feasible, future work should stratify evaluation by sub-variety.

\paragraph{Licensing and attribution}
We rely on publicly released corpora whose licensing permits research use. However, downstream redistribution of trained models may require additional license checks, especially for subsets with non-commercial clauses.

\paragraph{Human evaluation}
Our HES is based on three prompts and few raters. While OES mitigates BLEU’s brittleness, the human portion is small-scale and subject to annotator bias. Future work should increase rater diversity and adopt MQM-style guidelines.

\paragraph{Generalizability}
We investigate five pairs; results may not extrapolate to language families with different morphosyntax (e.g., polysynthetic or agglutinative languages with scarce English bitext). Moreover, we fine-tune encoder–decoder transformers; conclusions may differ for decoder-only or massively multilingual models.

\paragraph{Potential harms}
MT systems can propagate stereotypes or misrepresent sensitive content when trained on web-scale data. Although our datasets are relatively constrained, care should be taken before deploying to high-stakes settings (health, legal, immigration).

\section{Conclusion}

This paper has presented a comprehensive examination of machine translation transfer learning across diverse linguistic pairs, spanning multiple language families—Semitic, Bantu, Indo-European, and language isolates—and varying levels of resource availability. Using a controlled experimental framework with five distinct language pairs, we have provided empirical evidence that transfer learning is not confined to closely related languages within the same family, but can be applied successfully in cross-linguistic contexts. 

The primary conclusions of this work are:
\begin{enumerate}
    \item Transfer learning performance improves as the similarity between the parent and child languages increases; however, it remains viable even when languages are from different families.
    \item In this context, learning rates must remain within the range \(n \times 10^{-4}\) to avoid catastrophic failure, but exact values within this range do not significantly alter results.
    \item Effective fine-tuning typically requires a modest number of epochs (6–7), as the parent models already encode substantial linguistic knowledge.
    \item Weight decay has no statistically significant impact on performance in this setting.
    \item Optimal batch size varies with language similarity: medium batch sizes work best for more similar pairs, while smaller sizes are advantageous for less similar pairs that require finer-grained adaptation.
\end{enumerate}

\subsection*{Broader Implications}
These findings demonstrate that transfer learning can serve as a \emph{practical recipe} for low-resource MT: researchers and practitioners can begin with existing high-resource models, acquire only a few thousand parallel sentences, and achieve usable translation quality without massive computation. This has direct implications for endangered or minority languages, where data collection is costly and computational resources scarce. The results also highlight that typological similarity is not merely a linguistic curiosity but a measurable signal that can guide model selection and training strategy.

\subsection*{Future Directions}
Future work should broaden the typological range of studied pairs, including polysynthetic and agglutinative languages (e.g., Inuktitut, Quechua, Turkish) and languages with unique script systems. Another direction is \emph{cross-script transfer}, where transfer could help bridge Arabic-script and Latin-script varieties of the same language. Moreover, while this study focused on encoder–decoder transformers, it remains an open question whether similar patterns hold in decoder-only transformer models or in massively multilingual pre-trained models. Finally, multimodal extensions, such as speech translation, would test whether cross-linguistic transfer extends beyond text.

\subsection*{Ethical and Societal Considerations}
The results also highlight ethical issues. Leveraging standardized forms (e.g., Modern Standard Arabic) to improve colloquial varieties (e.g., Levantine) risks privileging dominant dialects while underrepresenting others. Similarly, Western Armenian datasets largely reflect diasporic communities, which may bias MT systems toward particular orthographic conventions. Careful dataset documentation, transparent licensing, and evaluation across regional varieties will be crucial in responsible deployment. In high-stakes domains—such as healthcare or legal contexts—MT errors can have significant consequences, underscoring the need for human-in-the-loop systems.

\subsection*{Closing Remarks}
In summary, this study provides empirical evidence that transfer learning is both \emph{effective and generalizable} across languages from distinct families. By demonstrating that only modest amounts of target-language data are needed, we offer a scalable path toward supporting under-resource languages in the global digital ecosystem. The hope is that these insights encourage both academic research and community-driven projects to apply transfer learning responsibly, ensuring that linguistic diversity is preserved and supported in the era of AI-driven communication.

\section{Acknowledgments}
This research was conducted as part of an exchange program at the Technical University of Darmstadt, under the supervision of Dr. Sabine Bartsch and her team in the Department of
Corpus- and Computational Linguistics, English Philology.

\bibliography{refs}

\end{document}